%% file: main.tex
\documentclass[journal,twoside]{IEEEtran}

\usepackage{enumitem}
\usepackage{lmodern}
\usepackage{listings}
\usepackage{array}
\usepackage{booktabs}
\usepackage{arydshln}
\usepackage{hhline}
\usepackage{ragged2e}
\usepackage{dirtytalk}
\usepackage{multicol}                 
\usepackage{setspace}
\usepackage{lipsum}
\usepackage{color}
\usepackage{multirow}
\usepackage{float}
\usepackage{csvsimple}
\usepackage{pifont}
\usepackage{algorithm}
\usepackage[noend]{algpseudocode}
\usepackage[caption=false,font=normalsize,labelfont=sf,textfont=sf]{subfig}
\usepackage{stfloats}
\usepackage{url}
\usepackage{verbatim}
\usepackage{graphicx}
\usepackage{cite}
\usepackage{orcidlink}
\usepackage{amsmath,amssymb,amsfonts}
\usepackage{hyperref}
\begin{document}
\title{Blockchain-Enabled Privacy-Preserving Second-Order Federated Edge Learning in Personalised Healthcare}
\author{Anum Nawaz\,\orcidlink{0000-0002-1148-0084}, Muhammad Irfan\,\orcidlink{0009-0005-4326-932X}, Xianjia Yu\,\orcidlink{0000-0002-9042-3730}, Hamad Aldawsari\,\orcidlink{0009-0008-5510-8605}, Rayan Hamza Alsisi\,\orcidlink{0000-0003-3448-9425}, Zhuo Zou\orcidlink{0000-0002-8546-1329}, Tomi Westerlund \,\orcidlink{0000-0002-1793-2694} (Senior IEEE Member)
\thanks{ Anum Nawaz is working with Shanghai Key Laboratory of Intelligent Information Processing Lab, Fudan University, China and Turku Intelligent Embedded and Robotic Systems (TIERS) Lab, University of Turku, Finland (e-mail: 18110720163@fudan.edu.cn), Muhammad Irfan and Xianjia Yu is working with TIERS Lab, University of Turku, Finland. Hamad Aldawsari is working with Haql University College of University of Tabuk, Saudi Arabia and Rayan Hamza Alsisi is with Department of Electrical Engineering, Islamic University of Madinah, Saudi Arabia.}
\thanks{Zhuo Zou (Senior Member, IEEE) is a Professor with the School of Information Science and Technology, Fudan University, Shanghai, China. (e-mail: zhuo@fudan.edu.cn)}
\thanks{Tomi Westerlund (Senior Member, IEEE) is a Professor with Robotics and Autonomous Systems at the University of Turku. He leads the TIERS Lab (tiers.utu.fi), Faculty of Technology, University of Turku, Finland. (e-mail: tovewe@utu.fi)}
}
\maketitle
\begin{abstract}
Federated learning (FL) is increasingly recognised for addressing security and privacy concerns in traditional cloud-centric machine learning (ML), particularly within personalised health monitoring such as wearable devices. By enabling global model training through localised policies, FL allows resource-constrained wearables to operate independently. However, conventional first-order FL approaches face several challenges in personalised model training due to the heterogeneous non-independent and identically distributed (non-iid) data by each individual's unique physiology and usage patterns. Recently, second-order FL approaches maintain the stability and consistency of non-iid datasets while improving personalised model training. This study proposes and develops a verifiable and auditable optimised second-order FL framework \textit{BFEL (blockchain enhanced
federated edge learning)} based on optimised \textit{FedCurv} for personalised healthcare systems. \textit{FedCurv} incorporates information about the importance of each parameter to each client's task (through fisher information matrix) which helps to preserve client-specific knowledge and reduce model drift during aggregation. Moreover, it minimizes communication rounds required to achieve a target precision convergence for each client device while effectively managing personalised training on non-iid and heterogeneous data. The incorporation of ethereum-based model aggregation ensures trust, verifiability, and auditability while public key encryption enhances privacy and security. Experimental results of federated CNNs and MLPs utilizing mnist, cifar-10, and PathMnist demonstrate framework's high efficiency, scalability, suitability for edge deployment on wearables, and significant reduction in communication cost.

\end{abstract}
\input{sections/1.Introduction}
\input{sections/2.ProposedModel}
\input{sections/3.Experiments}
\input{sections/4.PerformanceEvaluations}

\input{sections/5.Discussion_and_Analysis}
\bibliographystyle{IEEEtran}
\bibliography{bibliography}
\end{document}

%% file: sections/1.Introduction.tex
\section{Introduction}
Traditional ML methodologies necessitate training on data consolidated within a single data repository, which may be either centralised or distributed~\cite{simko2024reproducibility}. This paradigm requires raw data from multiple participants to be transmitted to a centralised aggregator server. However, aggregating data from multiple stakeholders in healthcare systems poses significant challenges, particularly concerning security, compromising data owners’ privacy, and possibly exposing sensitive health data and high latency rate~\cite{liu2021machine,ali2024novel}. 
Within the distributed ML paradigm, two primary frameworks exist: data centre-based distributed ML and cross-device FL. The former utilizes optimised computing nodes, data shuffling, and high-bandwidth communication networks, whereas the latter operates on a large number of resource-constrained devices with limited computational, storage, and communication capabilities~\cite{abdulrahman2020survey}. 

Federated edge learning (FEL) has emerged as a distributed ML paradigm that mitigates privacy concerns by facilitating collaborative model training while ensuring that data from wearables remains decentralised on the devices. FEL bridges these two paradigms, possessing computational and storage capacities comparable to data centre-based ML while sharing communication constraints with cross-device FL due to physical distance between devices, multi-hop transmissions, and diverse communication mediums. It offers a decentralised alternative, enabling multiple wearables to collaboratively train a model without sharing raw data, where each participant trains a model locally, and shares only the model parameters.
Some of the widely used first-order FL approaches are FedAvg, FedSgd, FedAdam, and FedYogi. While these methodologies preserve the confidentiality of the sensitive data, the resulting shared model parameters remain vulnerable to confidentiality breaches during aggregation and dissemination.
Despite its several advantages, FL faces two fundamental challenges: (i)managing hardware and data heterogeneity across a diverse fleet of consumer devices and (ii) handling real-world data that are often non-independently and identically distributed (non-iid) as physiological patterns are unique to each individual~\cite{zhu2021federated,abbasi2021eeg}. Specifically, the delivery of personalised FL services stands crucial in wearable technologies advancement because patients manifest individual health profiles with specific requirements.
Gradient based first-order approaches are widely used in FL to preserve the confidentiality of sensitive local data. However, this protection does not extend to the model parameters themselves, which remain vulnerable to confidentiality breaches during aggregation and dissemination. General training datasets have limited representation of particular classes or behaviors because specific data points are spread sparsely throughout the dataset. ~\cite{javeed2023federated}.

However, second-order FL methods provide better suitability for varied, heterogeneous wearable data present on different consumer clients. It allows better customisation of local models thus enhancing their value for personalised healthcare systems. Such adaptations provide potential benefits of privacy and security for consumer electronics while preserving personalization (i.e. higher model adaption variability) such as natural gradient descent~\cite{qi2024federated} and quasi-newton method~\cite{hamidi2025distributed}.

Nevertheless, the distributed implementation of second-order methodologies in a traditional manner remains challenging due to their reliance on inverse matrix-vector product computations, which introduce complexity in determining the descent direction. Addressing these challenges, it is crucial to develop efficient and scalable second-order optimization techniques tailored to heterogeneous FL environments. Second-order optimization methods offer a key advantage by incorporating curvature information of the loss function, thus improving personalised training processes.

Recently, some adaptations incorporate second-order curvature information using the fisher information matrix (FIM) for better convergence~\cite{jhunjhunwala2023towards}. Moreover, integration of distributed ledger technologies in FL approaches significantly improves systems auditability and transparency in data-sensitive health ecosystems to build reliable and unbiased intelligent systems~\cite{zhu2023blockchain}.

This study aims to optimize a distributed optimization algorithm \textit{FedCurv} that minimizes communication rounds from resource-limited wearables required to achieve a target precision for convergence while effectively managing personalised training on non-iid and heterogeneous data across consumer edge devices. 
The main contributions of this article:

\begin{enumerate}[label=\roman*).]
    \item A novel privacy-preserving \underline{b}lockchain enhanced \underline{f}ederated \underline{e}dge \underline{l}earning (\textit{BFEL}) framework, which maintains auditability, verifiability, availability and ensures privacy protection in edge-FL environments.
    \item We proposed optimised implementation of \textit{FedCurv} for heterogeneous medical data to improve real-time personalised health monitoring and prediction for wearable devices.
    \item Blockchain-based secure aggregation mechanisms to ensure trust, tamper-proof model updates, decentralised auditability, and accountability.
    \item Security, scalability, and correctness analysis of our proposed scheme, \textit{BFEL} demonstrate high performance and model utility while maintaining privacy.
\end{enumerate}

The remainder of this article is organised as follows. Section II presents the background and motivation, Section III details the proposed BFEL framework, and Section IV describes the experimental setup. Section V discusses the results, and Section VI concludes the paper with future research directions.

\begin{figure}[t]
{\includegraphics[width=\columnwidth]{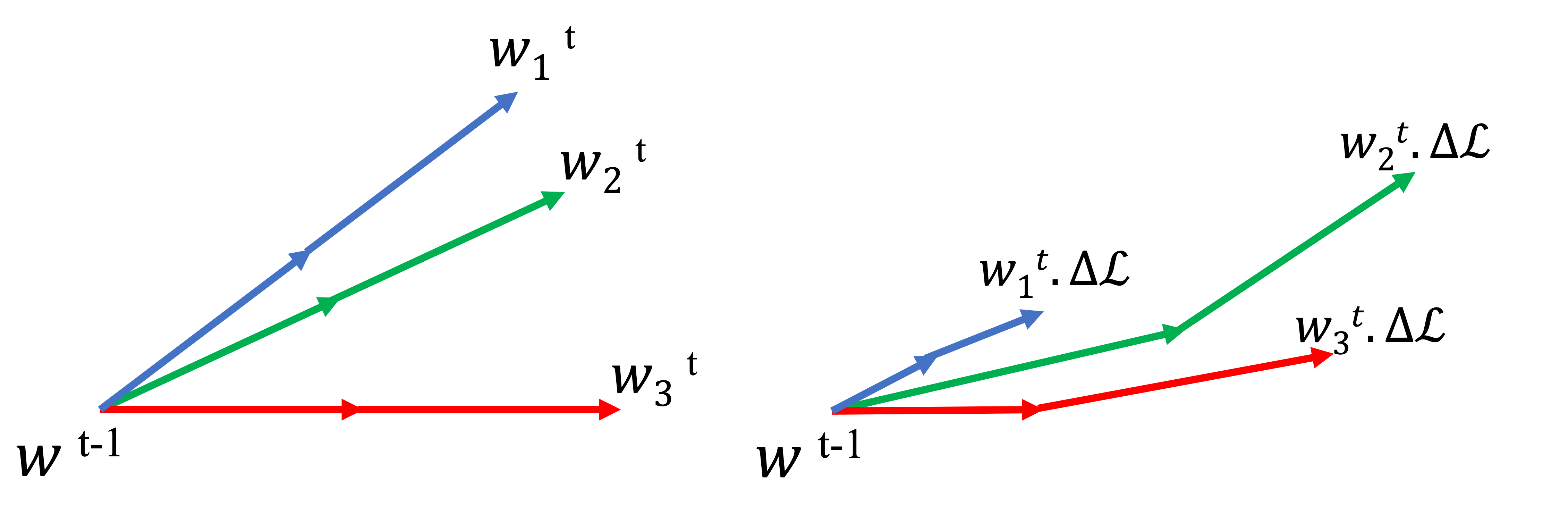}}
\caption{Left: Weight divergence in \textit{FedAvg} due to data heterogeneity. Right: \textit{FedCurv} induces a regularization parameter to minimize divergence to update weights according to less critical parameters
}
\label{fig:fedavgvsfedcurv}
\end{figure}

\section{Related Work}
Recent advancements in AI, particularly in the domain of DL within ML, have significantly contributed to the enhancement of smart and personalised healthcare ecosystems~\cite{bolhasani2021deep}. While the expansion of training datasets has been a key driver of progress, it simultaneously heightens the risk of privacy violations~\cite{lopez2023comprehensive}. Research indicates that such violations occur through privacy attacks which can extract sensitive information from the training data. This progress is largely driven by the expansion of training datasets. However, increased data volume also heightens the risk of privacy violations~\cite{lopez2023comprehensive}. Studies indicate that privacy attacks targeting DL models can lead to the inadvertent leakage of sensitive training datasets. Such privacy concerns present a substantial barrier to the continued advancement and deployment of DL technologies~\cite{otoum2024machine}. Centralised aggregation of private data for ML models poses significant challenges, especially in terms of security, privacy, and confidentiality~\cite{chen2022machine}. Consequently, retaining control over data without external dissemination to prioritize the security and privacy of data owners is an essential requirement for healthcare applications, motivating the creation of FL methods. Auditability and verifiability are essential components for establishing trustworthiness in FL. Several studies proposed blockchain-empowered FL approaches to enhance transparency, accountability, verifiability and the independent validation of FL processes~\cite{zhu2023blockchain}.

Achieving trustworthy FL systems requires mechanisms for auditability and verifiability. Numerous studies have suggested combining FL with blockchain technology to improve transparency, accountability, verifiability, and independent auditing of FL operations~\cite{zhu2023blockchain}. For example, some authors designed a blockchain-based trusted execution environment to protect local training data and introduced multi-signature verification for global models to strengthen auditability~\cite{kalapaaking2023blockchain}. Others have created traceable, transparent, and auditable supply chain solutions by optimizing data batching with Hyperledger Sawtooth~\cite{nawaz2024hyperledger}. Additional work includes smart contract-based local training policy enforcement and integrity verification for trained models~\cite{kalapaaking2023smart}, as well as peer-to-peer data sharing platforms that use blockchain to return data ownership rights to the original producers~\cite{nawaz2020edge}.

In FL research, first-order federated edge learning techniques, which rely solely on gradient information, are frequently utilised due to their robustness in distributed settings and minimal local computational requirements~\cite{li2025joint}. In comparison, second-order methods incorporate both gradient and curvature information, thereby facilitating improved descent direction selection and significantly accelerating convergence. This acceleration reduces the number of communication rounds required to achieve convergence, making second-order methods particularly advantageous in heterogeneous FL environments.

Both continual learning and FL approaches employ diverse strategies to address challenges like catastrophic forgetting, task interference, and communication efficiency~\cite{ma2022continual}. Elastic weight consolidation (EWC)~\cite{wen2018overcoming} mitigates catastrophic forgetting by restricting parameter updates critical to previous tasks using FIM, ensuring solutions compatible with both old and new tasks. In contrast, Incremental moment matching (IMM)~\cite{lee2017overcoming} models the posterior distribution of parameters for multiple tasks as a mixture of gaussians to harmonize task-specific knowledge. Stable SGD~\cite{jin2020stochastic} enhances performance by dynamically adjusting hyperparameters and incrementally reducing the learning rate upon encountering new tasks. For FL, \textit{FedCurv}~\cite{shoham2019overcoming} adapts a modified EWC framework to minimize disparities between client models during collaborative training as depicted in figure~\ref{fig:fedavgvsfedcurv}. Recent advancements~\cite{ayeelyan2025federated} further refine aggregation mechanisms using Bayesian non-parametric methods to improve model alignment across heterogeneous data sources. Communication overhead, a persistent hurdle in federated systems, is addressed~\cite{ren2025advances}, which employs layer-wise aggregation where shallow layers are updated frequently, while deeper layers are consolidated only in the final stages of training loops, significantly reducing bandwidth demands. Authors in~\cite{han2024fedsecurity} presents FedSecurity, a tool that simplifies testing security attacks and defenses in FL. It saves time by handling the underlying setup, allowing researchers to easily experiment with different models, datasets, and protection methods. In a recent study~\cite{wu2024fedbiot}, large language models are trained to work with private data at edge level, allowing owners to collaboratively train the model while utilising parameters from base model. Together, these methods collectively advance the robustness and scalability of learning systems in dynamic, distributed environments.

%% file: sections/2.ProposedModel.tex
\section{Privacy preserving \textit{BFEL} Service Framework}

\begin{figure*}[t]
\centering
\includegraphics[width=.9525\textwidth]{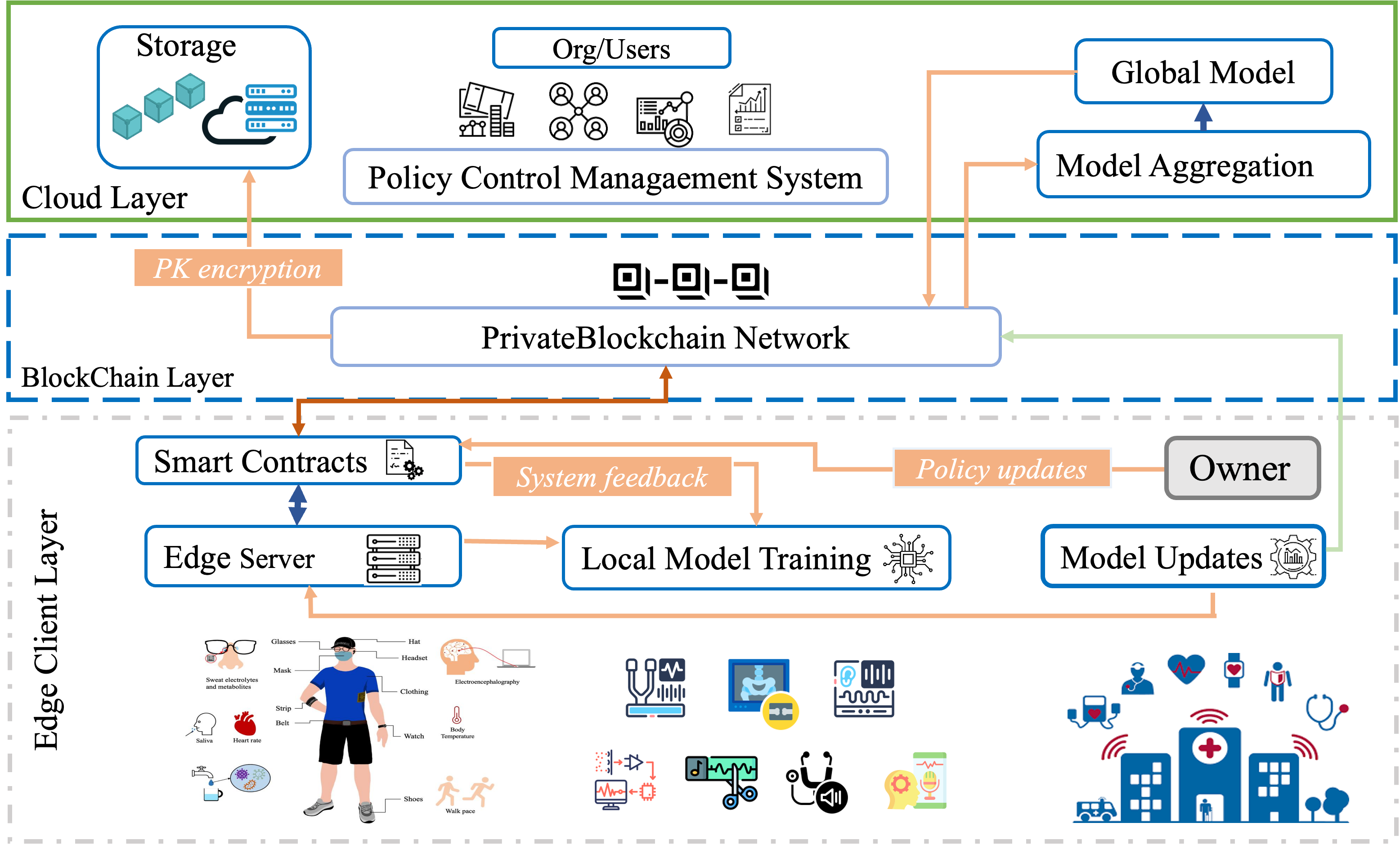}
 \caption{Hierarchical diagram depicting a sequential picture of complete network}
 \label{fig:frameworkk}
\end{figure*}

This section outlines the architectural flow of a proposed privacy-preserving \underline{b}lockchain enhanced \underline{f}ederated \underline{e}dge \underline{l}earning (\textit{BFEL}) framework, which maintains auditability, verifiability, availability and ensures privacy protection in edge-FL environments.
 To address the challenges of heterogeneous data in smart personalised healthcare devices, we propose an optimised second-order federated learning algorithm named federated curvature (\textit{FedCurv}). \textit{FedCurv} algorithm is built on elastic weight consolidation (EWC) to prevent catastrophic forgetting across edge devices and incorporates second-order information from the fisher information matrix (FIM) to preserve critical model parameters during the training process. The system-level flow of the proposed \textit{BFEL} framework is illustrated in figure~\ref{fig:frameworkk}.

\subsection{Edge Client Layer}

\begin{figure}[t]
\includegraphics[width=0.98\columnwidth]{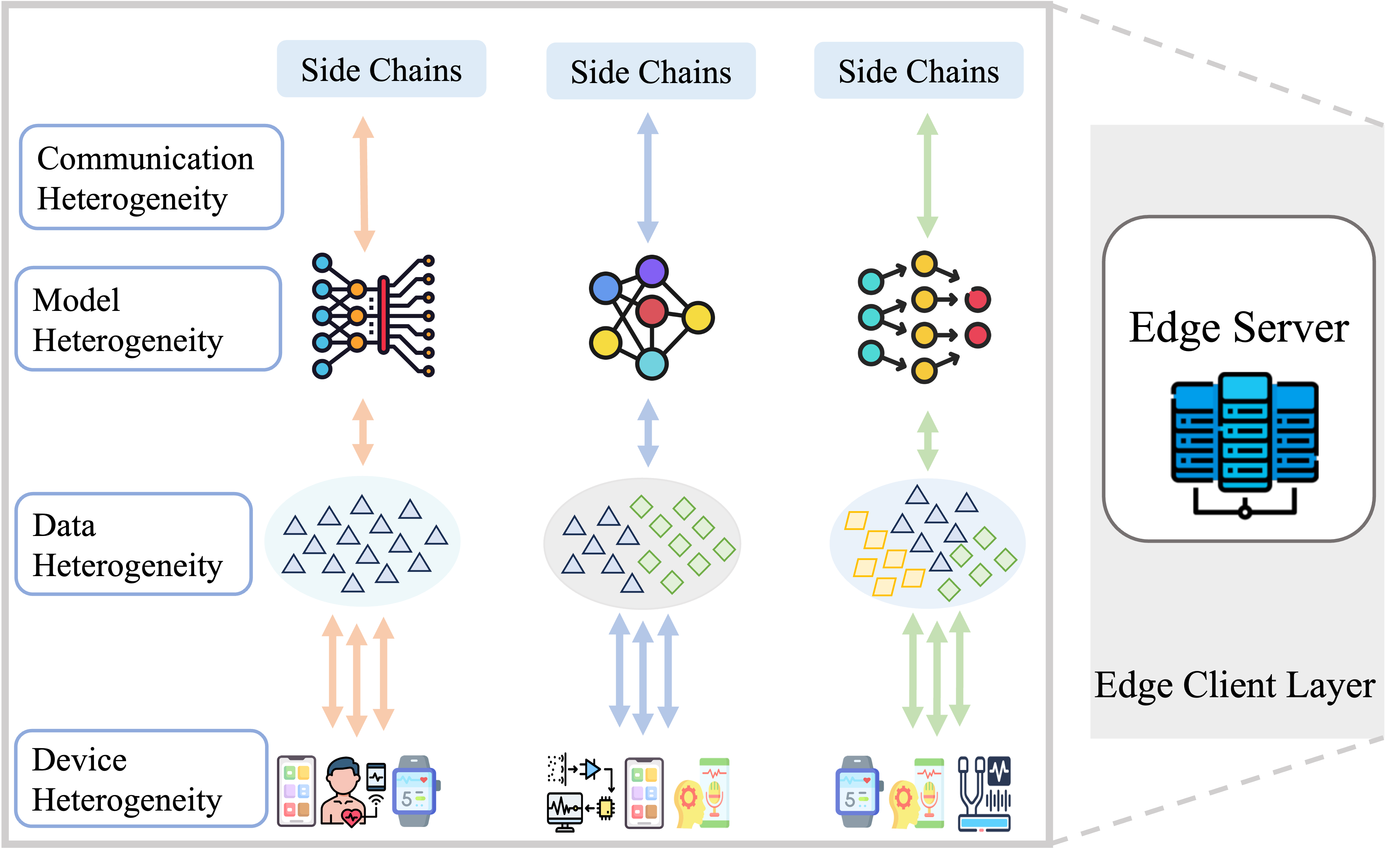}
 \caption{Edge Client Layer}
 \label{fig:framework1}
\end{figure}

The proposed architecture, shown in figure~\ref{fig:framework1}, comprises a multi-tiered system beginning with the edge clients layer. This foundational tier consists of edge devices, such as wearable technologies.
 Data sources encompass three primary categories: patients utilizing personal health monitoring devices, hospital electronic health record (EHR) systems, and diagnostic datasets from clinical or laboratory settings. In a network of $n$ clients denoted by $k=\left\{k_1, k_2, \ldots, k_n\right\}$ each hosting $m$ IoT devices for data acquisition, $D=\left\{d_1, d_2, \ldots, d_m\right\}$, we consider. The service run by each IoT device $d_j \in D$, referred to as the collection service ($S_{\text {col }}$) is responsible for gathering data (e.g. EEG signals).

Each client $k$ receives the global model parameters, denoted by $\theta_{\text{global}}$, from the server. Afterwards, each client computes the local FIM using the received global model parameters. Specifically, a diagonal approximation of the local FIM, $F_k$, is computed based on the client’s local dataset $\mathcal{D}_k$. Each diagonal element $F_k[i]$ is calculated as the average squared gradient of the log-likelihood of the data with respect to the global model parameters.

\begin{equation}
 F_k[i] = \frac{1}{|\mathcal{D}_k|} \sum_{(x, y) \in \mathcal{D}_k} \left( \nabla_{\theta_i} \log p(y|x; \theta_{\text{global}}) \right)^2   
\end{equation}

Subsequently, local training is performed using curvature regularization. The objective is to minimize a regularised loss function $\mathcal{L}_k(\theta)$, which combines the standard local loss with a curvature penalty term. The penalty term involves the squared difference between local and global parameters weighted by the diagonal FIM. The gradient of the regularised loss is then computed as the sum of the gradient of the local loss and a regularization term scaled by a factor $\lambda$. The model parameters are updated using stochastic gradient descent (SGD) for $E$ local epochs based on the gradient of this regularised loss. $F_k[i]$ corresponds to diagonal entry of $F_k$ for parameter $i$.

\begin{equation}
\nabla \log p(y|x; \theta)
\end{equation}
Above equation describes the gradient of the log-likelihood with respect to parameters. Compute local training with curvature regularization to minimize the regularised loss:

\begin{equation}
\mathcal{L}_k(\theta) = \underbrace{\text{Local Loss}}_{\mathcal{L}_{\text{LocalLoss}}} + \underbrace{2 (\theta - \theta_{\text{global}})^T \cdot \text{diag}(F_k) \cdot (\theta - \theta_{\text{global}})}_{\text{Curvature Penalty}}
\end{equation}

where $\lambda$ corresponds to regularization strength. Gradient calculation is computed as:

\begin{equation}
  \nabla \mathcal{L}_k = \nabla \mathcal{L}_{\text{LocalLoss}} + \lambda \cdot \text{diag}(F_k)(\theta - \theta_{\text{global}})
\end{equation}

Following this, update $\theta$ via SGD for $E$ epochs:

\begin{equation}
  \theta_{\text{local}} = \theta_{\text{global}} - \eta_{\text{local}} \cdot \nabla \mathcal{L}_k
\end{equation}

After local training, compute the gradient for server by computing the gradient of the \textbf{original loss} (without regularization) at $\theta_{\text{local}}$:
\begin{equation}
g_k = \nabla \mathcal{L}_{\text{LocalLoss}}(\theta_{\text{local}})
\end{equation}

In post-training, clients propagate their local model parameters $\theta_{\text{local}}$, curvature data and $g_k$ gradients to the aggregation server through blockchain. 
 This approach inherently safeguards data privacy by ensuring sensitive information remains within its original jurisdiction, serving as the primary defence against privacy breaches in sensitive healthcare applications. These updates are systematically broadcast across a private blockchain network, ensuring auditability and transparency.

\subsection{Blockchain Layer}
The blockchain layer serves as a foundational component of the proposed system, providing an immutable and transparent ledger to enhance security, accountability, and compliance in healthcare ecosystems. By design, the blockchain records all transactions including consent management, data access requests, and model updates in a tamper-evident manner, thereby ensuring full traceability across the network. 

Specifically, smart contracts are utilised as patient consent parameters, such as permissible data usage and authorized entities. These self-executing agreements autonomously enforce predefined policies, which eliminates manual oversight and reduces the risk of unauthorized access. Each step of the aggregation workflow such as server update requests, participant submissions, and the generation of the global model is cryptographically hashed and immutably logged on the blockchain, enabling independent verification of procedural integrity.  

The bloackchain broadcasts, stored the finalised global model parameters in an encrypted form using public-key infrastructure $pki$, ensuring that only authorised entities with corresponding decryption keys can access sensitive insights. Subsequent training rounds are initiated by broadcasting these encrypted parameters via the blockchain to aggregation servers, thereby maintaining synchronization while preserving security. To further reinforce access control, blockchain mediated authentication ensures that researchers' access requests trigger smart contracts. These contracts validate permissions against patient-defined policies before any approval is granted. All such interactions, including consent modifications and data retrieval attempts, are permanently recorded on the ledger, creating auditable trails for regulatory review. This architecture not only ensures adherence to standards such as health insurance portability and accountability act (HIPAA)~\cite{HIPA} and general data protection regulation (GDPR)~\cite{GDPR} but also empowers patients with visibility into data usage through transparent audit logs. By integrating decentralised consensus mechanisms, cryptographic encryption, and automated policy enforcement, the system establishes a robust framework for privacy preserving collaboration, balancing operational transparency with uncompromising data security.

\subsection{Cloud Server Layer}

Cloud servers are deployed virtually over the network and working as aggregation servers, collecting locally computed $\theta_{\text{local}}$ and gradients $g_k$ from each client device $k$. Compute aggregated curvature $F_{\text{global}}$ to get important scores (FIMs) from all clients and average gradients $g_{\text{global}}$. 

\begin{equation}
  F_{\text{global}} = \frac{1}{K} \sum_{k=1}^{K} F_k
\end{equation}

\begin{equation}
  g_{\text{global}} = \frac{1}{K} \sum_{k=1}^{K} g_k
\end{equation}

Secondly, compute inverse FIM for diagonal $F_{\text{global}}$

\begin{equation}
F_{\text{global}}^{-1}[i] = \frac{1}{F_{\text{global}}[i] + \epsilon} \quad
\end{equation}

Inverse FIM scores invert the average importance scores, which ensures priority of important parameters (with large curvature values) gets smaller updates. It helps the model learn from new data without forgetting important knowledge from previous clients.

Following this, server clouds update the global model by applying curvature-scaled update:
\begin{equation}
\theta_{\text{global}}^{\text{new}} = \theta_{\text{global}} - \eta_{\text{global}} \cdot \left( F_{\text{global}}^{-1} \odot g_{\text{global}} \right)
\end{equation}

Finally, calculated $\theta_{\text{global}}^{\text{new}}$  global updates are broadcast to the clients through the blockchain broadcasts using smart contracts.
\subsection{Communication Model}

The system employs a layered communication model and protocol architecture designed to ensure privacy-preserving, secure, and auditable operations across asynchronous phases. Data collection is initiated by sensors and client devices, which acquire raw physiological metrics such as blood glucose levels and electrocardiogram signals. Https/gRPC encrypts FL client-server communications, while mqtt optimizes lightweight data transfer from IoT/wearables to the client servers.

Our proposed access scheme is used along with the widely used elliptic curve integrated encryption scheme (ECIES), child key derivation function (ckd), and the elliptic curve digital signature algorithm (ECDSA) are cryptographic techniques to ensure secure data storage and communication. ECIES works by independently deriving a bulk encryption key and a mac key from a common secret. The data is first encrypted under a symmetric cipher, and then the ciphertext is authenticated under an authentication scheme. Finally, the common secret is encrypted under the public part of a public-private key pair. The ckd function is used for managing data in batches, child key derivation functions are used in hierarchical deterministic wallets (hd wallets). It helps in generating a tree of keys from a single master key, which can be very useful for managing multiple keys securely and systematically. ECDSA is a digital signature algorithm that is used for secure key sharing for both data and communication. The ECDSA ensures that the data and communication are coming from the stated sender (authenticity), have not been altered in transit (integrity), and repudiation by the sender can be disputed (non-repudiation). Implementing ECIES, ckd, and ECDSA in this proposed system pipeline provides a robust framework that ensures secure data storage and communication. The ECIES offers a strong encryption scheme for data protection, the ckd provides an efficient way to manage data in batches, and the ECDSA guarantees secure key sharing and data authenticity.

Following this, local model training at client device level is completed using \textit{FedCurv}, model local weights $F_k$, $g_k$ are transmitted through private ethereum network, ensuring the confidentiality of individual client $k$ data weights, preventing re-identification. Subsequently, processed data is transmitted to the model aggregation servers via a blockchain network, where aggregated global models are cryptographically hashed and immutably logged on the blockchain ledger. Transaction validation occurs through a distributed consensus protocol, ensuring integrity. In this patient-centric framework, patients configure data sharing permissions via a portal, triggering ethereum based smart contracts, while researchers and clinicians submit queries via blockchain transactions, which undergo automated authorization checks. Approved requests retrieve models or insights from secure decentralised storage systems such as IPFS, with all access events (identity, timestamp, purpose) permanently recorded on-chain for auditability. Role-based access control (RBAC) policies are programmatically enforced through smart contracts. 

%% file: sections/3.Experiments.tex
\section{Experimental setup}
This section details the experimental setup designed to evaluate the performance of the proposed \textit{BFEL} framework. To address data heterogeneity across clients, we employ the optimised \textit{FedCurv} algorithm as our core federated learning method and compare its performance against the baseline algorithm, \textit{FedAvg}.
The FL network is configured with each participant performing 10 local training rounds per global round, and 20 global aggregation rounds.
We utilise three benchmark datasets: mnist, cifar-10, and medmnist. mnist and cifar-10, as standard benchmarks in neural network research, while medmnist specifically its pathmnist subset, derived from 2D image classification. On the client side(\ref{alg:client_fedcurv}), each client device performs local training while incorporating a curvature-aware regularization term. This regularization penalizes deviations from the global model based on the estimated importance of each parameter, thereby preserving critical knowledge and improving model stability. On the server side (\ref{alg:server_fedcurv}), the curvature matrices and gradients collected from the clients are aggregated. The server then applies a curvature-scaled update to the global model, ensuring that parameter adjustments are inversely proportional to their estimated importance. This approach enables more efficient and robust federated optimization in non-iid settings characterised by unbalanced data distributions ~\cite{kalsoom2025deep}.
A private ethereum blockchain network is utilised as a service layer to demonstrate its capability to uphold auditability, verifiability, and availability while ensuring privacy preservation in edge-FL settings. 

\subsection{Parameter Settings}
To evaluate the proposed framework under realistic conditions, we utilise three distinct benchmark datasets, each selected to inquire different capabilities of the system. The mnist dataset, a standard benchmark for image classification, includes 70,000 images (28×28 pixels), split into 60,000 training and 10,000 test samples, providing a baseline for foundational algorithm validation. For a more clinically relevant and challenging benchmark, we employ the medmnist subclass pathmnist (28×28 pixels), a more challenging alternative, comprises 35,000  training images and 8000 testing images in non-iid, enabling evaluation of model generalizability in scenarios with higher intra-class variability. Thirdly, cifar-10, a widely used dataset for object recognition, offers 60,000 color images (32×32 pixels). Together, these datasets simulate real-world edge-FL challenges, such as decentralised computation and heterogeneous data privacy requirements, while validating the framework’s ability to balance transparency, security, and efficiency in privacy-sensitive environments. 

 \subsection{Network Configuration:}
This study utilizes two widely recognised neural architectures: a multi-layer perceptron (MLP) and a convolutional neural network (CNN). These models served as the foundational deep learning frameworks for training classification systems within a FL setup, simulating server client training scenarios. The experiments aimed to assess how effectively each algorithm handles non-iid data and preserves knowledge across distributed medical and non-medical imaging tasks.

For the image-based datasets, mnist and pathmnist, the CNN architecture consists of two convolutional layers, each followed by a max-pooling layer for spatial downsampling, and concludes with two fully connected layers.
To accommodate the RGB input channels of the cifar-10 dataset, the architecture is modified while retaining the core structure. Both configurations employ the SGD optimizer with a learning rate of 0.001.

\subsection{Hardware and Software Configuration}
We utilise raspberry pi3 model B+ minicomputers as edge device servers (manager nodes of side chains) and lightweight nodes are implemented using STM32F427 development boards (low-power ARM cortex M3,M4 and M7 processors), which are used for high-speed implementation of asymmetric cryptographic algorithms. ECDSA is used to generate public and private keys, and device authentication mechanisms. The STMicroelectronics x-cube-cryptolib library is utilised to implement several standard cryptographic algorithms with the ARM cortex-M series processors. System components are developed using golang, solidity to write smart contracts and deployed using remix IDE, and metamask handles concurrent transactions and interactions with third-party cloud services. A PoS consensus ensures block validation, while the gossip protocol enables fast, resilient message propagation and node synchronization. Figure \ref{fig:managernodes} depicts the system’s initialised nodes.
 
\begin{algorithm}
\caption{FedCurv Server Side Computation}
\label{alg:server_fedcurv}
\begin{algorithmic}[1]
\Require Initial model $\theta^0_{\text{global}}$, total rounds $T$, number of clients $K$, server learning rate $\eta_{\text{global}}$, numerical stability $\epsilon$
\Ensure Trained global model $\theta^T_{\text{global}}$
\State Initialize global model 
\For{round $t = 1$ \textbf{to} $T$}
    \State Broadcast $\theta_{\text{global}}$ to all participating clients    
    \State Collect client updates  
    \State Aggregate Fisher Information Matrices       
    \State Aggregate gradients   
    \State Compute inverse Fisher information   
    \State Update global model
\EndFor
\State \Return $\theta_{\text{global}}$
\end{algorithmic}
\end{algorithm}

\begin{algorithm}
\caption{FedCurv Client Side Computation}
\label{alg:client_fedcurv}
\begin{algorithmic}[2]
\Require Global model parameters $\theta_{\text{global}}$, local dataset $\mathcal{D}_k$, regularization strength $\lambda$, local epochs $E$, learning rate $\eta_{\text{local}}$
\Ensure Updated Fisher Information Matrix $F_k$, \\ gradient $g_k$
\State Client Update   
\State Compute Fisher Information Matrix (FIM)
\State Initialize local model   
\For{epoch $= 1$ \textbf{to} $E$}
    \For{\text{each batch } $(x_b, y_b) \in \mathcal{D}_k$}
            \State Compute regularised loss  
            \State Compute gradient
            \State Update local model
        \EndFor
    \EndFor   
    \State Compute Server Gradient
    \State \Return $\{F_k, g_k\}$
\end{algorithmic}
\end{algorithm}

%% file: sections/4.PerformanceEvaluations.tex
\section{Performance Evaluations}

To validate the effectiveness of \textit{BFEL}, we conducted experiments across varying task sequence configurations using \textit{FedCurv} and compared its performance against established baselines using \textit{FedAvg}. The results are analysed through multiple perspectives, offering distinct insights. Code is available publicly at our github account\footnote{https://github.com/AnumNawazKahloon/FedCurv}.

\subsection{Federated Simulation Results}
\begin{figure*}[t]
\centering
\includegraphics[width=.99\textwidth]{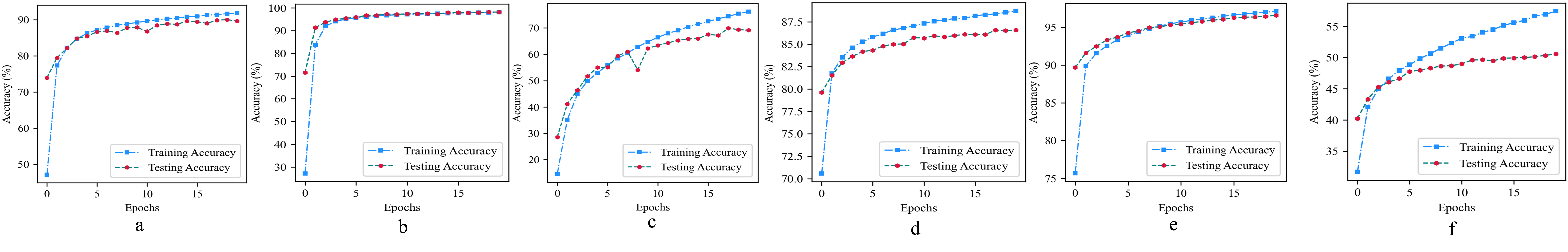}
 \caption{\textit{FedCurv} for Base (a) CNN PathMnist (b) CNN Mnist  (c) CNN Cifar (d) MLP PathMnist (e) MLP Mnist  (f) MLP Cifar}
 \label{fig:FedcurvBase}
\end{figure*}

\begin{figure*}[t]
\centering
\includegraphics[width=.99\textwidth]{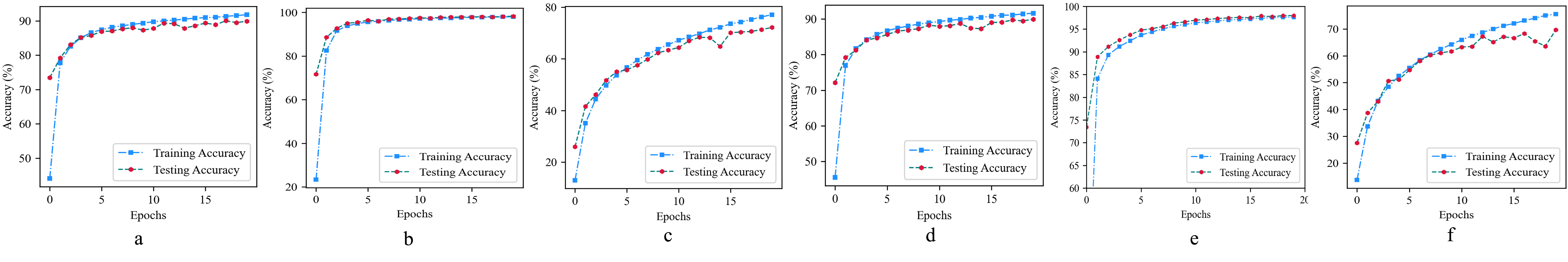}
 \caption{\textit{FedAvg} for Base (a) CNN PathMnist (b) CNN Mnist  (c) CNN Cifar (d) MLP PathMnist (e) MLP Mnist  (f) MLP Cifar}
 \label{fig:FedAvgBase}
\end{figure*}

From the epochs per round standpoint, accuracy consistently improves as the number of local training epochs increases across all settings and algorithms. This trend depicts the close alignment of local optima with global optima, making extended local training within each round advantageous when maintaining a fixed number of communication rounds. Accuracy of \textit{FedAvg} increases sharply after each epoch as compared to \textit{FedCurv} at the base level depicted in figure~\ref{fig:FedAvgBase} and ~\ref{fig:FedcurvBase} as well as edge client level, which shows the minimum divergence of results after each round in \textit{FedCurv}, makes it more consistent and according to the previous weights of edge client devices. 

In an overall comparison of performance, \textit{FedCurv} designed primarily to address non-iid data challenges in FL, surprisingly outperforms \textit{FedAvg} even in uniform data settings. Notably, \textit{FedCurv} often achieves superior accuracy after 100 rounds, indicating a potential need for extended training phases to reach convergence compared to \textit{FedAvg}. Regarding communication efficiency, reducing the frequency of communication rounds while keeping the total number of training epochs constant yields better model performance, implying that less frequent parameter exchanges may enhance learning stability or optimization.

The experimental setup employed the \textit{SGD} optimizer with an adaptive learning rate decay, reduced by a factor of three after every five epochs, along with a mini-batch size of 20 wit 20 rounds per task, and 1 epoch per round. 
For the MLP configuration, adjustments included a reduced mini-batch size of 10 and an initial learning rate of $1\mathrm{e}{-4}$, while client sampling fractions of 0.25 and 0.05 were applied at each round. Hyperparameters $\lambda1$ and $\lambda2$ were fixed at $[1\mathrm{e}{-1}, 4\mathrm{e}{-1}]$ and 100, respectively, across all experiments. These findings collectively highlight the interplay between local training intensity, algorithmic robustness, and communication strategies in FL frameworks.

\subsubsection{Client side training}
\begin{figure*}[t]
\centering
\includegraphics[width=.99\textwidth]{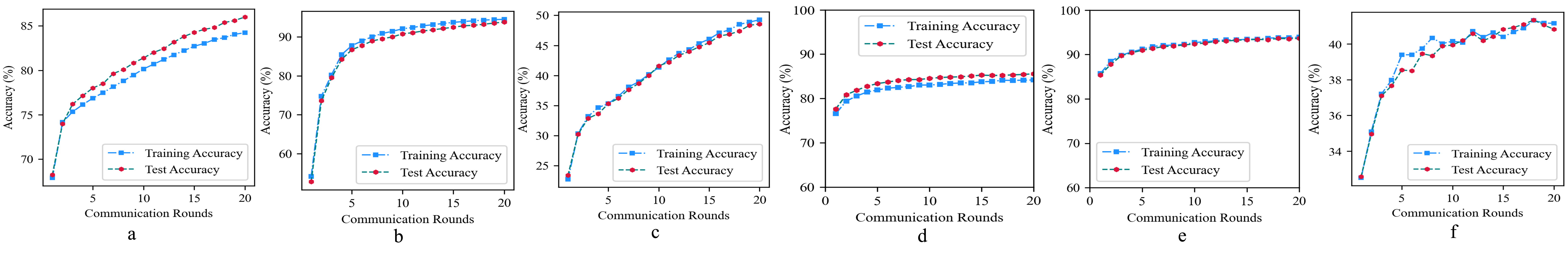}
 \caption{\textit{FedCurv} for Federated Non-iid (a) CNN PathMnist (b) CNN Mnist  (c) CNN Cifar (d) MLP PathMnist (e) MLP Mnist  (f) MLP Cifar}
\label{fig:fedcurvFed}
\end{figure*}

\begin{figure*}[t]
\centering
\includegraphics[width=.99\textwidth]{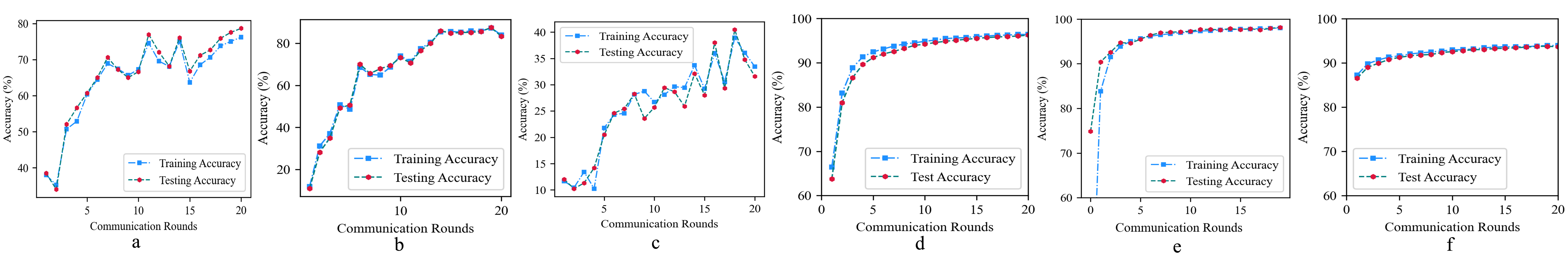}
 \caption{\textit{FedAvg} for Federated (a) Non-iid Fed CNN PathMnist (b) Non-iid Fed CNN Mnist  (c) Non-iid Fed CNN Cifar, (d) IId Fed CNN PathMnist (e) IId Fed CNN Mnist (f) IId Fed CNN Cifar }
 \label{fig:fedAvgFed}
\end{figure*}

In the edge client side, federated learning experiments, \textit{FedCurv} demonstrated consistent performance as shown in ~\ref{fig:fedcurvFed} over \textit{FedAvg}~\ref{fig:fedAvgFed}, particularly in non-iid settings with CNN models. On the pathmnist dataset using CNN, \textit{FedCurv} achieved significantly better accuracy of around ~86\% and exhibited stable convergence, while \textit{FedAvg} struggled with fluctuations and lower accuracy of ~75\%. Similarly, for CNN on MNIST, \textit{FedCurv} outperformed \textit{FedAvg} with a final test accuracy around 95\%, compared to \textit{FedAvg}’s noisier convergence and lower accuracy of ~85\%. The difference was significant on cifar-10 with CNN, where \textit{FedCurv} showed a gradual increase to ~50\% accuracy, outperforming \textit{FedAvg}, which displayed highly unstable learning and plateaued around 40\%. After each epoch round, \textit{FedCurv} maintains low divergence in non-iid settings. 

\subsubsection{Server side training}
However, during server side base experiments across the mnist, pathmnist, and cifar-10 datasets. \textit{FedAvg} consistently outperformed \textit{FedCurv} in terms of final testing accuracy and convergence behavior. For CNN based experiments, \textit{FedAvg} achieved slightly higher accuracies across all datasets. On the pathmnist dataset, \textit{FedAvg} reached around 90\% accuracy, while \textit{FedCurv} trailed slightly at approximately 88\%. A similar trend was observed for mnist, where \textit{FedAvg} achieved about 99\% accuracy compared to \textit{FedCurv}’s 98\%. On the more complex cifar-10 dataset, \textit{FedAvg} demonstrated better generalization, achieving approximately 72\% accuracy, while \textit{FedCurv} lagged behind at around 68\%.

\subsection{Layer 2 Ethereum Implementation}
The proposed system implements a second layer ethereum network is implemented using a polygon sidechain structure, which operates on e proof of stake (PoS) consensus. It consists of two main components, a primary blockchain layer hosted on cloud servers and a sub-blockchain network of multiple sidechains. Sidechains are deployed on individual edge clients to facilitate localised model training in \textit{BFEL}.
Within each sub-blockchain, the corresponding edge client functions as a miner node, responsible for performing local model training and updating local model weights based on the global model parameters after each training epoch. Subsequently, each edge client transmits its locally trained results to the main blockchain layer, where global model parameters are aggregated. 


\begin{figure}
\includegraphics[width=\columnwidth]{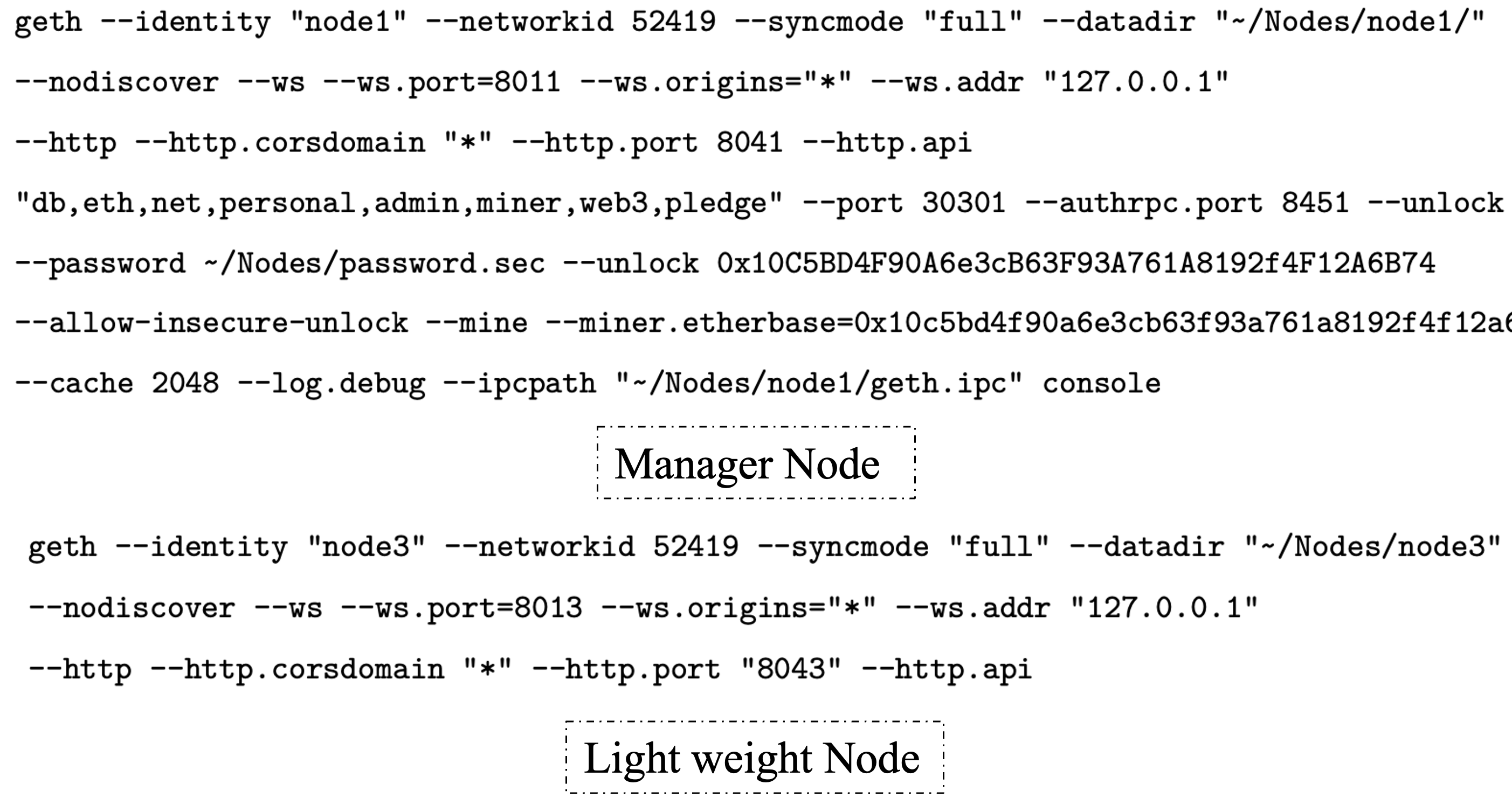}
\caption{Manager nodes and lightweight nodes after layer2 ehtereum initialization}
 \label{fig:managernodes}
\end{figure}

The edge clients are also tasked with collecting and packaging model training outputs (treated as transactions) into candidate blocks, which are then published to their respective private blockchain network. The main blockchain layer is operated by cloud servers configured as a consortium blockchain. Each registered user organization such as hospitals or governmental agencies within the consortium is permitted to establish its own private sidechain. Furthermore, to reduce potential latency caused by complex task assignment mechanisms, transaction packaging responsibilities in both private and consortium blockchains are delegated to designated manager nodes.
\begin{center}
\begin{table*}
\footnotesize
\centering
\caption{Average \textit{FedAvg} versus \textit{FedCurv} edge client-side accuracy and average server-side base learning accuracy results}
\begin{tabular}[t]{@{}p{2.0cm} p{1.8cm} p{1.7cm} p{1.7cm} p{1.7cm} p{1.7cm}@{}}
\toprule
 \textbf{Dataset} & \textbf{Model}  & \textbf{client Acc \textit{FedAvg} }  & \textbf{client Acc \textit{FedCurv}} & \textbf{Base Acc \textit{FedAvg}} & \textbf{Base Acc \textit{FedCurv}} 
 \\ \midrule \midrule
PathMNIST & CNN & \textasciitilde75\% & \textasciitilde86\% & \textasciitilde90\% & \textasciitilde88\%
\\
MNIST & CNN & \textasciitilde85\% & \textasciitilde95\% & \textasciitilde99\% & \textasciitilde97\%
\\ 
CIFAR-10  & CNN & \textasciitilde40\% & \textasciitilde50\% & \textasciitilde72\% & \textasciitilde68\%
\\ 
PathMNIST & MLP & \textasciitilde95\% & \textasciitilde85\% & \textasciitilde91\% & \textasciitilde88\%
\\ 
MNIST & MLP & \textasciitilde94\% & \textasciitilde94\% & \textasciitilde97\% & \textasciitilde95\%
\\ 
CIFAR-10 & MLP & \textasciitilde45\% & \textasciitilde42\% & \textasciitilde70\% & \textasciitilde50\%
\\ \bottomrule
\end{tabular}
\end{table*}
\end{center}
\subsection{Security Analysis}
We have conducted a performance evaluation of three different types of STM32F427 M series processors within the framework of asymmetric cryptography. We utilised the x-cube-cryptolib library to implement the ECDSA. To ascertain the statistical error of the results obtained over the number of executions, we calculated the mean, standard deviation, and standard error using the appropriate equations.

\begin{figure}
\includegraphics[width=\columnwidth]{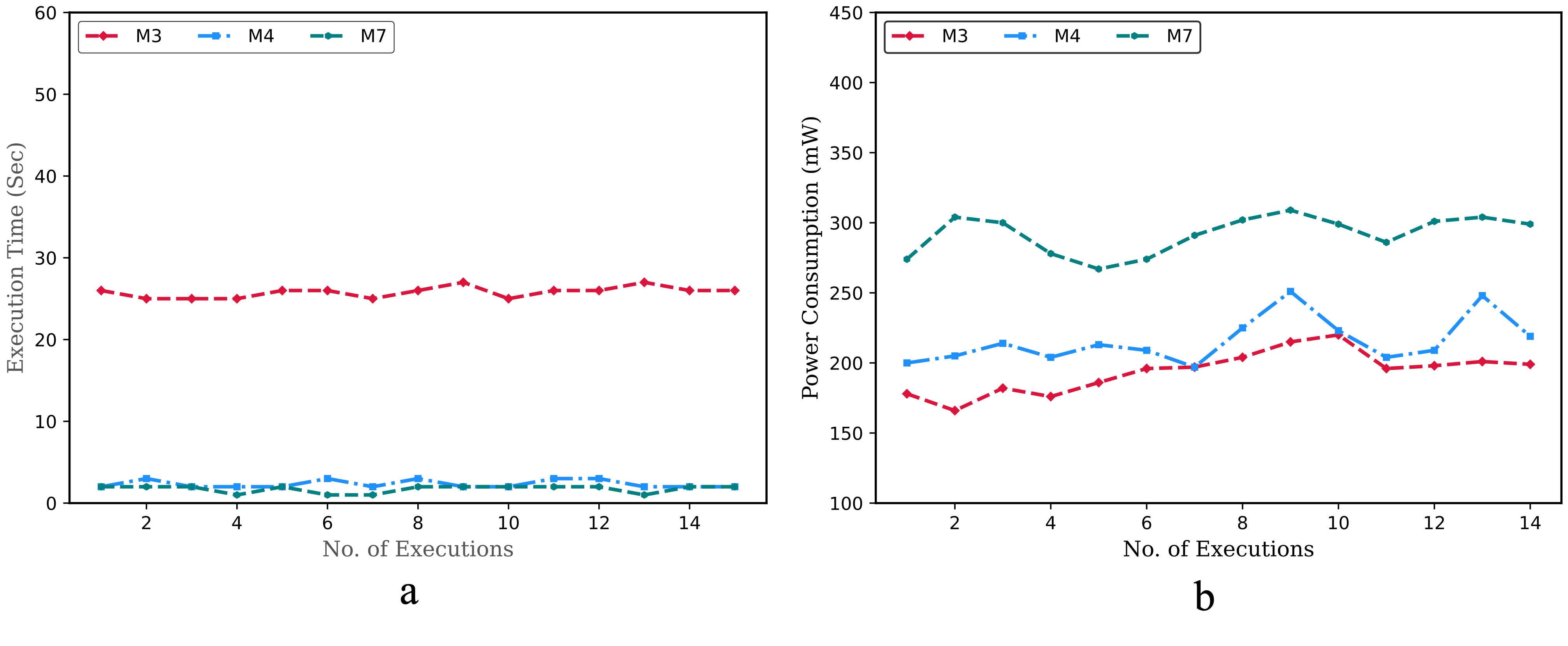}
\caption{(a) Execution time and (b) Average power consumption,  during ECDSA implementation using STM32F427 M series processors}
 \label{fig:ECDSAexecutiontime}
\end{figure}

The execution time was determined, which encompasses the total time required for key generation, encryption, and decryption using ECDSA. To identify the optimal execution time of ECDSA, we examined the records of different numbers of executions for each processor. The execution time for each processor for ECDSA is visually represented in figure~\ref{fig:ECDSAexecutiontime} subsection (a). For a comprehensive analysis, the execution time was calculated in terms of mean, standard deviation, and standard error for each processor. The estimated execution time of ECDSA for processor M3 is 26.352 s $\pm$ 0.002s, and the execution time for M4 processor is 1.451s $\pm$ 0.007s and 1.167s $\pm$ 0.002s for M7. Based on the results, the average execution times of M3 processors are 17.253 seconds, while the execution times for processors M4 and M7 are 1.462 seconds and 1.156 seconds, respectively. The data suggests that the M4 and M7 processors exhibit superior performance in executing ECDSA. These time measurements facilitate easy planning and adjustments to determine the delay tolerance in the network. 

This study evaluates power consumption, a key microcontroller parameter, during the execution of cryptographic algorithms. Power was measured by calculating current consumption from the voltage drop across a shunt resistor (R) using Ohm's law. The current consumption of the processors was calculated using ohm’s law. To determine the average power consumption, ECDSA was executed for 15 runs and a comparison of power consumption is presented in figure~\ref{fig:ECDSAexecutiontime}(b). The average power consumption by M3 and M4 was $\pm$ 200mW, whereas M7 used an average of $\pm$ 290mW. Results indicate the superior performance of M4 cortex microcontrollers are best fit while consuming fewer resources. 

A deterministic wallet is used for ckd functions to determine a child's key from a parent's key. Using this technique, each batch of data to be encrypted in the device is given a unique secret key ~\cite{antonopoulos2017mastering}. The 512-bit hash is calculated according to the parent's public key (public and private keys are 256 bits) and the desired child index. It is impossible to deduce the original parent key from the n$th$-child key because of the one-way hashing used in the process. This process appears to generate random numbers due to the additions \textit{modulo n}.

\subsection{Scalability Analysis}
Client based DLTs offer several offer significant benefits, but also face inherent limitations in terms of scalability limitations that restrict number of parallel processes. Nevertheless, the \textit{BFEL} handles this challenge using the side chains concept, a gossip protocol, and PoS consensus. FoBSim simulation tool\footnote{https://github.com/sed-szeged/FobSim} is utilised to check the scalability of the proposed model. Manager nodes ranging from 5 to 500 were used to check the performance matrix of the proposed model and measure the total time required to complete the request procedure at clients level devices versus at the cloud layer. 

\begin{figure}[t]
{\includegraphics[width=\columnwidth]{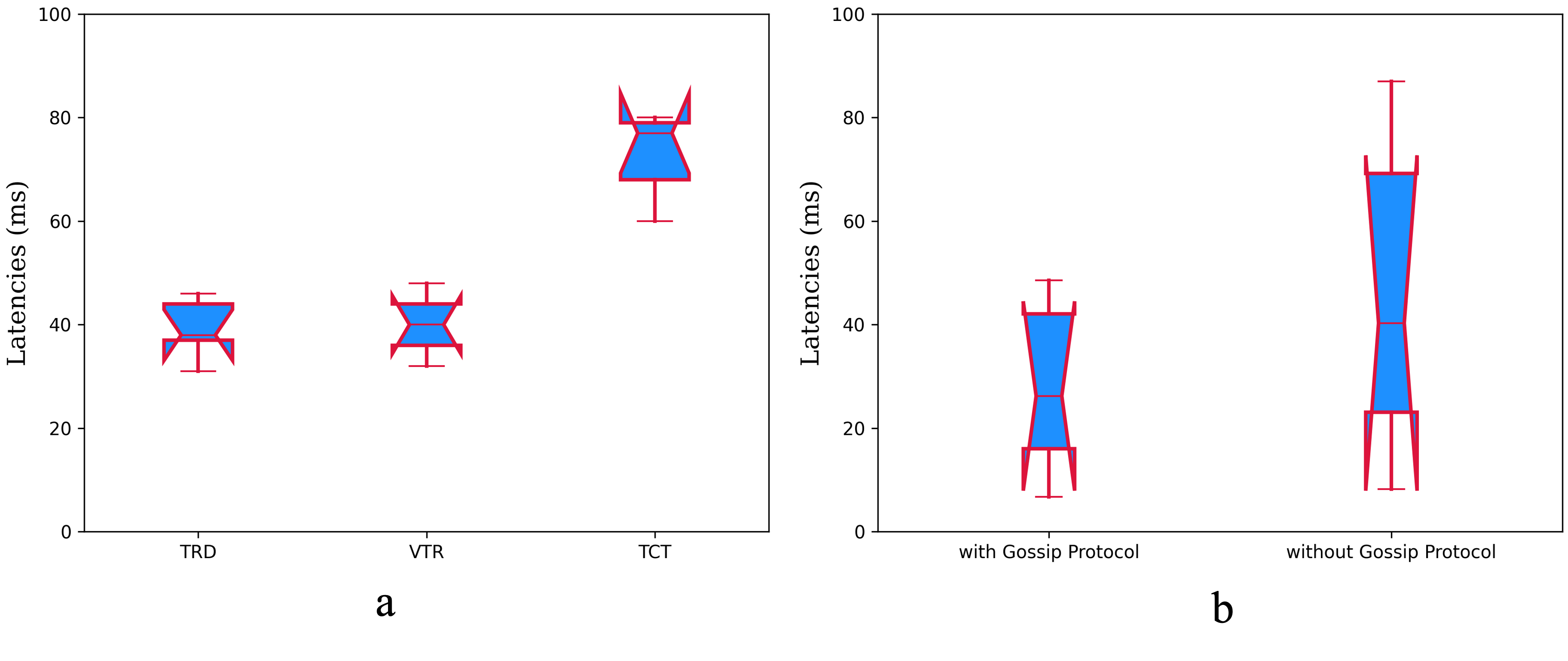}}
\caption{ (a) Latencies to retrieve mata data (TRD), transaction validation time (VTR) needed for single transaction request, and time required to confirm one transaction (TCT), (b)Impact of concurrent transactions on end-to-end transaction latency using gossip protocol versus without gossip protocol}
\label{fig:TCTTRD}
\end{figure}

The client measures and divides the time needed to complete a transaction into three sections: time to retrieve data (TRD), time for checking the transaction (VTR), and time for confirming the transaction (TCT). Figure~\ref{fig:TCTTRD}(a) presents the results of the measurements. Based on the available resources, it is impressive that TRD requires only 34.6 milliseconds on average, VTR 36 milliseconds on average, and TCT 73.6 milliseconds on average. Additionally, it is essential to note that TCT also relies on the network, which in this experiment was adversely affected by our shared Wi-Fi's slow response time, causing the time to be extended overall. 

\begin{figure}[htbp]
\centerline{\includegraphics[width=\columnwidth]{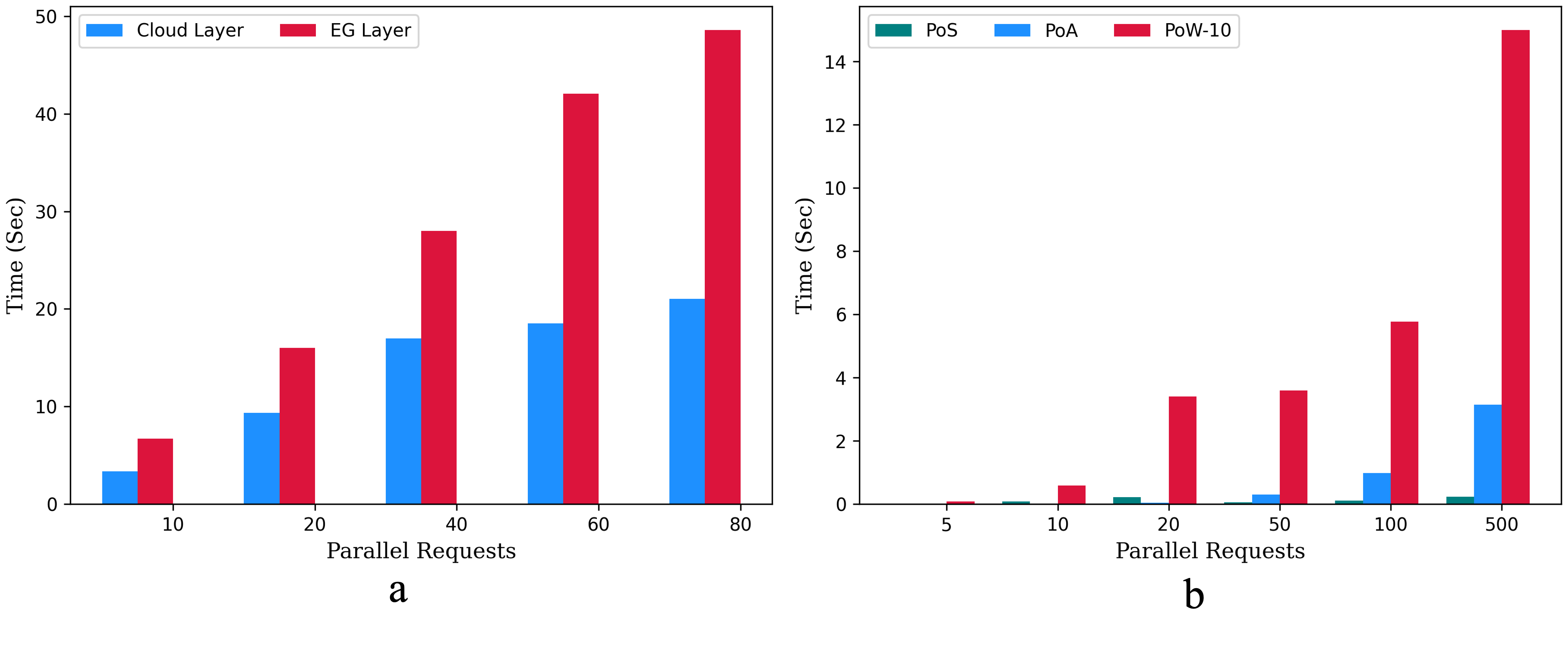}}
\caption{(a) Impact of concurrent transactions on end-to-end transaction latency on \textit{BFEL} versus cloud services, (b) Average Block Confirmation Time using PoS, PoW-10, and PoA}
\label{fig:totalelapsedtime}
\end{figure}

The gossip protocols are integrated to improve the scalability of the proposed network model and \textit{public key infrasture} (PKI) is utilised for cryptographically signing model updates and transactions. Results in ~\ref{fig:TCTTRD}(b) demonstrates the clear reduction in total elapsed time compared to transactions processed without gossip protocol. Figure~\ref{fig:totalelapsedtime}(a) shows that the cloud layer utilises around double the time as compared to edge clients to complete transaction requests during concurrent transactions starting from 5 to 100 transactions at a time. This architecture integrates end-to-end privacy pipelines, tamper evident logging, and granular access controls, ensuring regulatory compliance and transparency across all operational phases.

\textit{end-to-end delay = request initialization by interested buyer + time to retrieve metadata\\ + response time by manager nodes + time to confirm one transaction}

 Figure~\ref{fig:totalelapsedtime}(a) illustrates that increasing concurrent transaction requests leads to increased end-to-end delays. This study's results demonstrate the proposed model's effectiveness for autonomously implementing data sharing in information-critical systems. Using this trustless structure, data trade becomes more reliable and transparent. We have concluded that single-board computers can act as data and transaction managers, with no need for third-party cloud services, as the necessary computation makes space for other edge services and data processing processes to run simultaneously. However, a parallel number of transactions will cause a significant delay.  

Block confirmation time was also measured using different numbers of manager nodes, and the average time is illustrated in the figure for proof of stake versus two famous consensus algorithms PoW and PoA ~\ref{fig:totalelapsedtime}(b). \textit{BFEL} does not present a scalability issue due to the fact that it works in a private P2P network that can be segmented into side chains. It does not require edge clients to process many requests per minute. The manager nodes must check whether new requests for data have been received after every time $T$. If it has a queue of requests, the manager nodes will respond to each request one at a time.

%% file: sections/5.Discussion_and_Analysis.tex
\section{Discussion and Analysis}

In this study, we propose and develop a personalised healthcare system \textit{BFEL} based on second order FEL. Second-order FL methods bring a substantial benefit for personalised training through their capability to use loss function curvature data to improve personalised training procedures during local data training.
The \textit{BFEL} framework integrates blockchain as a service layer to ensure data privacy, transparency and confidentiality of data owners while maintaining auditability, verifiability, availability, and robust security across FEL environments.
Upon model training completion, both global model and local parameters are securely stored on the blockchain and subsequently distributed to the FL edge client devices and cloud servers in accordance with the established policy. Smart contracts are utilised to broadcast policies while providing automated confirmation and distribution operations.

To enhance personalized health monitoring and prediction\cite{abbasi2024deep}, we employ an optimised \textit{FedCurv} algorithm. It minimizes the necessary communication rounds for training on non-iid and heterogeneously distributed data across edge client devices. Performance evaluations demonstrate better performance of \textit{FedCurv} and resistance capabilities than \textit{FedAvg} using CNNs and MLPs in non-iid conditions. Key evaluation metrics including throughput, accuracy, privacy, and scalability analysis. 

Analysis of the experimental results demonstrates that the proposed method preserves the throughtput and accuracy of the ML process while ensuring auditability and verifiability throughout the training and aggregation procedures. Privacy protection is achieved through second-order FEL using \textit{FedCurv} and public key encryption ECDSA and ECIES, while latency and throughput were evaluated by measuring communication transactions on a permissioned blockchain and compared against a benchmark model \textit{FedAvg}. The results highlight that \textit{BFEL} outperforms the benchmark by achieving enhanced privacy, accuracy and scalability. The results underscore better stability and accuracy of the proposed framework \textit{BFEL} in heterogeneous settings and non-iid environments where patient data varies widely.